\documentclass[10pt,twocolumn,letterpaper]{article}

\usepackage{iccv}  
\usepackage{amsmath} 
\usepackage{amssymb} 
\usepackage{graphicx} 
\usepackage{makecell}
\usepackage{multirow}
\usepackage{adjustbox}
\usepackage{enumitem}
\usepackage{tabularx}
\usepackage{algorithm}
\usepackage[english]{babel}
\usepackage{algorithmic}

\definecolor{cvprblue}{rgb}{0.21,0.49,0.74}
\usepackage[pagebackref,breaklinks,colorlinks,allcolors=cvprblue]{hyperref}
\usepackage{xspace}

\newcommand{\name}{SIMS\xspace}

\title{\name: Simulating Stylized Human-Scene Interactions\\ with Retrieval-Augmented Script Generation}

\author{Wenjia Wang$^{1}$ \quad \; Liang Pan$^{1,2}$ \quad \; Zhiyang Dou$^1$ \quad \;  Jidong Mei$^1$ \quad \;  Zhouyingcheng Liao$^{1}$ \quad \; \\ Yuke Lou$^1$ \quad \; Yifan Wu$^1$ \quad \; Lei Yang$^2$ \quad \;  Jingbo Wang$^{2\dagger}$ \quad \; Taku Komura$^{1\dagger}$ \\[1.5mm]
\normalsize $^1$ The University of Hong Kong \quad
\normalsize $^2$ Shanghai AI Laboratory \quad
}

\newcommand{\ours}{SIMS\xspace}
\newcommand{\walk}{Walk\xspace}
\newcommand{\lie}{Lie\xspace}
\newcommand{\idle}{Idle\xspace}
\newcommand{\carry}{Carry\xspace}
\newcommand{\reach}{Reach\xspace}
\newcommand{\sit}{Sit\xspace}
\newcommand{\getup}{GetUp\xspace}
\newcolumntype{L}[1]{>{\raggedright\arraybackslash}p{#1}}
\newcolumntype{C}[1]{>{\centering\arraybackslash}p{#1}}
\newcolumntype{R}[1]{>{\raggedleft\arraybackslash}p{#1}}

\usepackage{amsmath,amsfonts,bm}
\usepackage{pifont}

\def\eqref#1{equation~\ref{#1}}

\def\1{\bm{1}}

\def\rva{{\mathbf{a}}}

\def\rvc{{\mathbf{c}}}

\def\rvg{{\mathbf{g}}}
\def\rvh{{\mathbf{h}}}

\def\rvs{{\mathbf{s}}}

\def\rvz{{\mathbf{z}}}

\DeclareMathAlphabet{\mathsfit}{\encodingdefault}{\sfdefault}{m}{sl}
\SetMathAlphabet{\mathsfit}{bold}{\encodingdefault}{\sfdefault}{bx}{n}

\begin{document}

\twocolumn[{
\renewcommand\twocolumn[1][]{#1}
\maketitle
\begin{center}
    \centering
    \captionsetup{type=figure}
    \includegraphics[width=0.96\linewidth]{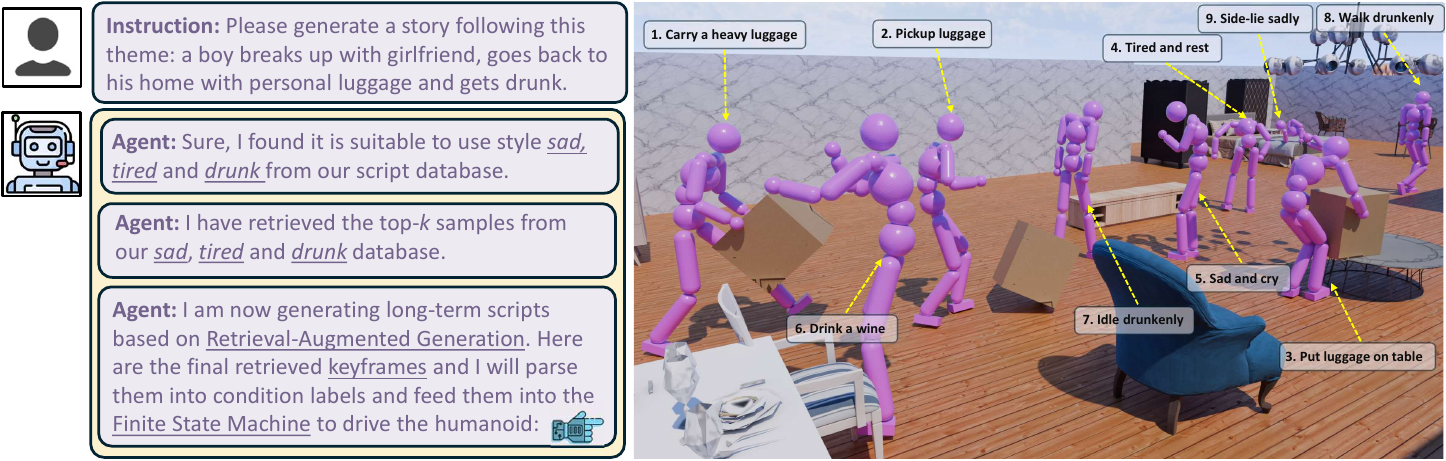}
    \captionof{figure}{
    \name enables physically simulated characters to 
    perform diverse skills within complex 3D scenes 
    given long-term daily narratives and scene inputs. 
    Our character could perform versatile skills, including Locomotions, Human Scene Interactions 
    and Dynamic Object Interactions 
    with diverse styles while accomplishing physically plausible contacts and obstacle avoidance. Left: a dialogue-based retrieval-augmented script generation process. Right: a skillful humanoid performing diverse stylized interactions in a 3D scene.}
   
    \label{fig:teaser}
\end{center}
}]

{
  \renewcommand{\thefootnote}%
    {\fnsymbol{footnote}}
  \footnotetext[1]{$\dagger$: equal advising.}
}

\begin{abstract}
Simulating stylized human-scene interactions (HSI) in physical environments is a challenging yet fascinating task. Prior works emphasize long-term execution but fall short in achieving both diverse style and physical plausibility. 
 To tackle this challenge, we introduce a novel hierarchical framework named \name that seamlessly bridges high-level script-driven intent with a low-level control policy, enabling more expressive and diverse human-scene interactions. Specifically, we employ Large Language Models with Retrieval-Augmented Generation~(RAG) to generate coherent and diverse long-form scripts, providing a rich foundation for motion planning. A versatile multi-condition physics-based control policy is also developed, which leverages text embeddings from the generated scripts to encode stylistic cues, simultaneously perceiving environmental geometries and accomplishing task goals. By integrating the retrieval-augmented script generation with the multi-condition controller, our approach provides a unified solution for generating stylized HSI motions. We further introduce a comprehensive planning dataset produced by RAG and a stylized motion dataset featuring diverse locomotions and interactions. Extensive experiments demonstrate \name's effectiveness in executing various tasks and generalizing across different scenarios, significantly outperforming previous methods. Project page: \href{https://wenjiawang0312.github.io/projects/sims/}{https://wenjiawang0312.github.io/projects/sims/}.

\end{abstract}   
\section{Introduction}
\label{sec:intro}
\begin{table*}
\vspace{-3mm}
\center
\adjustbox{max width=\linewidth}{
\begin{tabular}{l|C{75pt}|C{50pt}C{65pt}|C{50pt}C{55pt}C{75pt}|C{20pt}C{20pt}C{20pt}C{20pt}C{20pt}C{20pt}C{20pt}}
\toprule[1.5pt]
\multirow{2}{*}{\textbf{Method}} & \multirow{2}{*}{\textbf{Physical-Plausibe}}&
\multicolumn{2}{c|}{\textbf{Planner}}                                                              & \multicolumn{3}{c|}{\textbf{Controller}}                                                              & \multicolumn{7}{c}{\textbf{Incorporated Skills}}                                                                                          \\ 
&  & Automatic  & Style-Diversity & Text-Aware & Scene-Aware & Skill-Scalability & Walk & Sit & Lie & GetUp & Reach & Idle & Carry  \\ \midrule[1pt]
NSM\cite{nsm}&    \ding{55}       &    \ding{55}                     &         \ding{55}       &     \ding{55}    &        \ding{51}    &   \ding{55}   & \ding{51} &  \ding{51}  & \ding{55} & \ding{51}  & \ding{55} & \ding{51} & \ding{51} \\\midrule
SAMP\cite{samp} &       \ding{55}    &  \ding{55}               &    \ding{55}        &     \ding{55}    &    \ding{51}        &   \ding{55}   & \ding{51} &  \ding{51}  & \ding{55} & \ding{51}  & \ding{55} & \ding{55} & \ding{55}   \\\midrule
Humanise\cite{humanise} &  \ding{55} &          \ding{55}    &                 \ding{55}        &   \ding{51}      &         \ding{51}   & \ding{55} &  \ding{51}  & \ding{51} &\ding{51} &\ding{51} &\ding{55} & \ding{55} &\ding{55} \\\midrule
AffordMotion\cite{moveasyousay} &         \ding{55}  &       \ding{55}    &                  \ding{55}        &      \ding{51}    &  \ding{51}   &   \ding{55}   & \ding{51}   &  \ding{51}   &    \ding{51}   &   \ding{51}  & \ding{55} & \ding{55} & \ding{55}  \\ \midrule
TesMo\cite{tesmo}           &   \ding{55}         &      \ding{55}          &         \ding{55}   &            \ding{51}    &  \ding{51}  &  \ding{55}   &  \ding{51}    &  \ding{51}    &    \ding{51}    & \ding{51}     & \ding{55} & \ding{55} & \ding{55}    \\ \midrule[1.5pt]
InterScene\cite{interscene}          &      \ding{51}                                   &  \ding{51}     & \ding{55}  &  \ding{55}      &      \ding{51}            &  \ding{51}   &  \ding{51}  & \ding{51}    &  \ding{51}    &    \ding{51}  & \ding{55} & \ding{55} &\ding{55}  \\ \midrule
UniHSI\cite{unihsi}           &        \ding{51}                       &           \ding{51} &  \ding{55}       &      \ding{55}      &      \ding{51}       &   \ding{55} &  \ding{51}  & \ding{51}   &   \ding{51}   &    \ding{51} & \ding{51} &\ding{55}&\ding{55} \\ \midrule
\textbf{\ours(ours)}              &  \ding{51}   &   \ding{51}                 &      \ding{51}      &       \ding{51}           &    \ding{51} &   \ding{51} &  \ding{51}  & \ding{51}     &     \ding{51} & \ding{51} & \ding{51} & \ding{51}& \ding{51} \\ \bottomrule[1.5pt]
\end{tabular}}
\vspace{-2pt}
\caption{Comparision of Kinematics-Based(upper 5) and Physics-Based(lower 3) Long-term Human Scene Interaction methods.}
\label{tab:tasks}
\vspace{-8pt}
\end{table*}

Developing skillful characters with a broad repertoire of motor skills, such as walking, sitting, and reaching—while facilitating rich interactions with their environments has long been a desirable goal for animation, robotics, and VR/AR applications. In particular, achieving \textit{long-term, stylized, and physically plausible} interactions with diverse styles and intricate details is crucial for bringing characters and narratives to life.

Previous works~\cite{nsm, humanise, samp, wang2022towards, couch, zhao2023synthesizing} have explored long-term motion generation for kinematics-based human-scene interactions. However, they typically suffer from severe physical artifacts such as penetration and foot skating. To address these issues, recent studies~\cite{interphysics, interscene, unihsi, yuan2023physdiff, padl, sun2024prompt, hierarchical} have started incorporating physics simulators, i.e., ~\cite{isaacgym} to produce more physically plausible motions. Despite these advancements, the frameworks are limited to a small number of specific skills and task objectives, lacking diversity. %
Moreover, their planning results are often simplistic by following chronological lists~\cite{interscene,hierarchical} or focusing solely on contacts~\cite{unihsi}. This stands in contrast to real-world situations where 
body language in human motion and interactions directly convey a large number of \textbf{\textit{emotional or stylized}} states. For example, a person sitting on a chair with their head down and supporting it with their hands often conveys a sense of depression.

To address the aforementioned challenges, we propose a novel framework terms \ours, (\textbf{S}multating styl\textbf{I}zed hu\textbf{M}an \textbf{S}cene interactions). Specifically, \ours utilizes an LLM~\cite{gpt4} as a powerful high-level motion planner and physical policies as low-level controllers equipped with diverse motor skills. Inspired by Retrieval-Augmented Generation~\cite{RAG}, to generate semantically rich scripts, we develop a method of first creating a short script database and then retrieving and generating longer scripts. Each short script includes several keyframes detailing stylized interactions that the low-level control policy can effectively execute. We then retrieve the top-\textit{k} short scripts via the CLIP~\cite{CLIP} similarity between short script summaries and the user-provided story themes. Finally, we prompt the LLM to retrieve and generate stylized long-term scripts based on the short script inputs. 
Given the planned keyframes, a low-level control policy is employed to obtain the detailed body motions in the physical simulator, producing natural, diverse, and high-quality interactions. To ensure stylized motions are adaptable to various furniture shapes within a complex indoor environment, we propose a multi-condition control policy that is attuned to scene geometries, task goal observations, and text embeddings from the CLIP model~\cite{CLIP} for high-fidelity motion generation. 
Our multi-condition design not only facilitates effective scene perception but also captures fine-grained body movements, enabling a better grasp of stylized motor skills, i.e., the policy learns to perform more skills during imitation learning. Compared to previous policies~\cite{interphysics,interscene} that lack style control and UniHSI~\cite{unihsi}, which relies on accurate references, our approach supports flexible multi-condition control while mitigating mode collapse in AMP-based methods~\cite{amp}.
We incorporate a finite state machine (FSM) to manage multiple policies guided by specified keyframes, enabling the synthesis of physics-based animation that aligns with real-world distributions while improving scalability. To address the scarcity of motion data in the field of stylized motion generation, we collected and annotated captions and style labels from five existing motion capture datasets. Additionally, we capture a new dataset named ViconStyle to supplement the limitations in both the categories and quantity of stylized motion data.

We conduct an extensive evaluation of our method to validate its effectiveness. To provide a more comprehensive overview, we compare five SOTA kinematics-based~\cite{nsm,samp,humanise,moveasyousay,tesmo} and two physics-based~\cite{interscene,unihsi} long-term HSI methods with \ours to explain our task setting in \cref{tab:tasks}. Our method, \ours, surpasses existing approaches with a fully automatic framework that integrates style diversity, text awareness, scene awareness, and physics plausibility for realistic human-scene interactions. Unlike prior methods, it supports easy extension, ensuring scalability and adaptability. \ours also achieves the most comprehensive skill coverage, making it a state-of-the-art solution for versatile and controllable motion synthesis.
In summary, our contributions are threefold:
\begin{enumerate}
\item We propose a framework for physically simulated characters to perform stylized 3D interactions using RAG-based script generation and a multi-condition control policy that encodes style from text while adapting to the environment, featuring: \textit{(a) Stylized Control}: A script planner for coherent storytelling and a text-conditioned controller for expressive, style-consistent motion. \textit{(b) Automatic Generation}: A planner that generates executable keyframes from theme descriptions. \textit{(c) Scalability}: New skills and styles can be integrated by updating the script database and training a new policy.

\item We provide a comprehensive dataset of restructured motion clips with captions, emotional labels, and a short script database for stylized interactions.

\item Our method outperforms previous approaches across multiple metrics, achieving high-quality, diverse, and physically plausible long-term motion generation.
\end{enumerate}

\section{Related Works}
\label{sec:rw}

\noindent\paragraph{Kinematic-based Human Scene Interaction}
Synthesizing realistic human behavior has been a long-standing challenge. While most methods enhance the quality and diversity of humanoid movements \citep{motiondiffuse, motionclip, mdm, emdm, motiongpt, tlcontrol, cong2024laserhuman, lu2024scamo}, they often overlook scene interactions. Recently, there's been growing interest in integrating human-scene interactions, crucial for applications like embodied AI and virtual reality. Many previous approaches \citep{nsm, samp, wang2022towards, couch, humanise, trumans, moveasyousay, tesmo, cong2024laserhuman,zhang2024roam} rely on data-driven kinematic models~\cite{aios,zolly,tore, humanwild, disentangled} for static or dynamic interactions. However, these often lack physical plausibility, resulting in artifacts like penetration, floating, and sliding, and require additional post-processing, limiting real-time use.

\noindent\paragraph{Physics-based Human-Scene Interaction}
While previous physics-based animation approaches mainly focused on human motion alone~\cite{peng2022ase, amp,peng2018deepmimic, case, huang2025modskill}. InterPhys~\cite{interphysics} presents a framework extending AMP to include character and object dynamics, using a scene-conditioned discriminator for superior performance compared to previous methods. 
Additionally, InterScene~\cite{interscene} effectively synthesizes physically plausible long-term human motions in complex 3D scenes by decomposing interactions into Interacting and Navigating processes. This method uses reusable controllers trained in simple environments to generalize across diverse scenarios. 
With the development of LLMs, UniHSI~\cite{unihsi} introduces a unified framework for human-object interaction via language commands, featuring an LLM Planner and Unified Controller, which reduces training labor with LLM-generated plans. The effectiveness of this approach is evaluated using the ScenePlan dataset.

\noindent\paragraph{Comparison with Previous HSI Methods}
We compare five kinematics-based SOTA and two physics-based long-term HSI methods with SIMS to explain our task setting in \cref{tab:tasks}. NSM~\cite{nsm} and SAMP~\cite{samp} use goal positions for planning. Humanise~\cite{humanise}, AffordMotion~\cite{moveasyousay}, and TeSMo~\cite{tesmo} utilize text-based control for human motion, with the latter two leveraging textual annotations from datasets like HumanML3D~\cite{humanml3d}, enabling some details in motion expression. All five kinematics-based methods rely on continuous keyframe control, requiring frequent user input updates. In contrast, InterScene~\cite{interscene} automates control by setting long-term keyframes for FSM to switch skills, and  UniHSI~\cite{unihsi} applies long-term keyframes of body-object contacts. Our planning uses RAG to generate long-term scripts, and enable automation and diversity. For HSI skills, we focus on 2 locomotion skills: walk and idle, 4 common human scene interaction skills: sit, lie, get up, and touch, and 1 dynamic object interaction skill: carry. Regarding control extensibility, only InterScene and our approach allow training solely for new skills without retraining the entire controller. In Supp.Mat, we demonstrate how to easily involve new interaction skills with specific styles into our framework.

\begin{figure*}[ht]
    \centering
\includegraphics[width=0.96\linewidth]{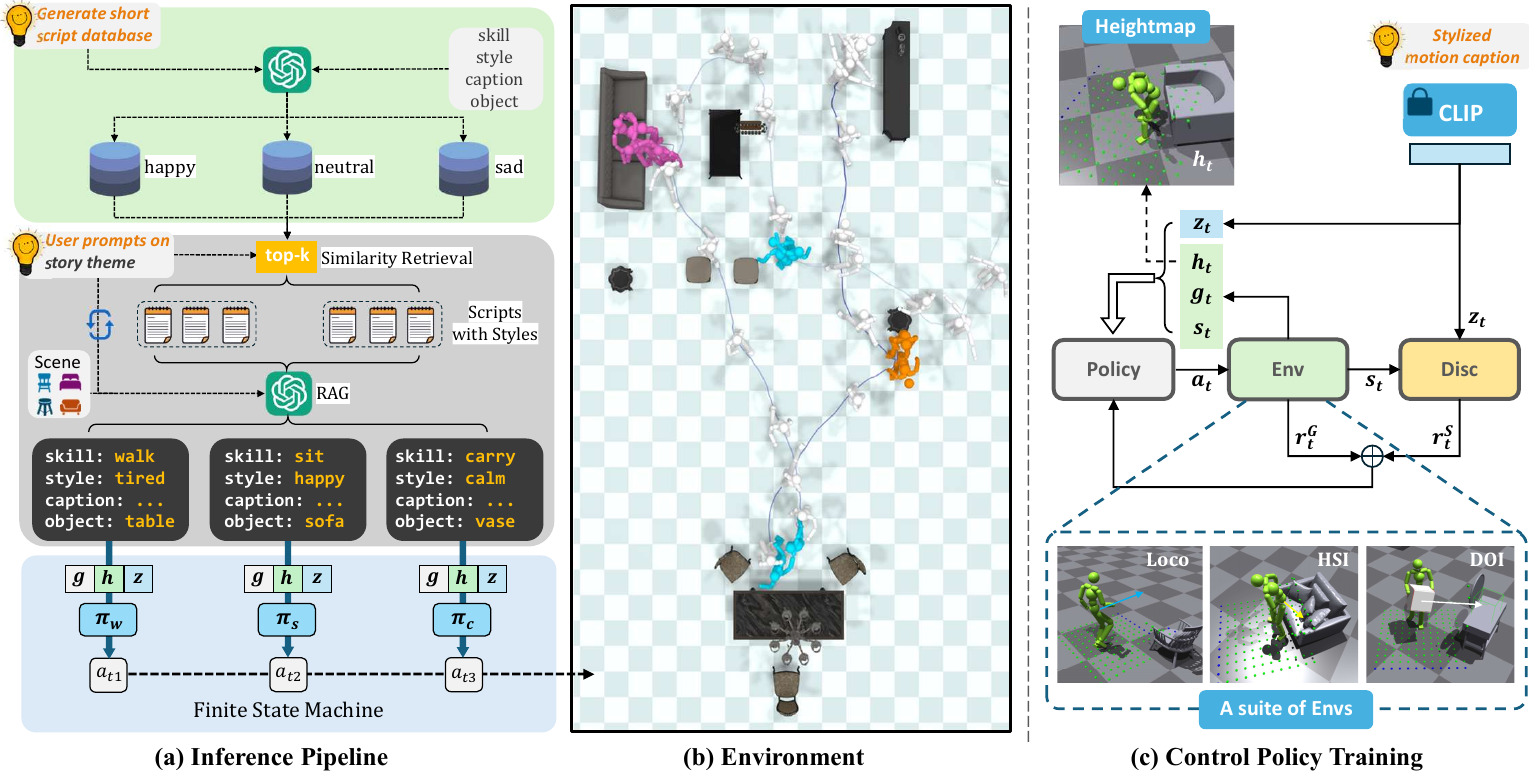}
    \caption{(a) Our main pipeline. We prompt LLMs to generate new short scripts following their emotion and interaction logic. The retrieval process includes 2 stages. We first retrieve the top-k short script with semantics similarity, then ask LLM to retrieve useful samples from the short scripts and concatenate them as a fluent long-term story.  
    In the Finite State Machine. We parse skills, captions,  and scene geometry from each keyframe into task goals, language embeddings, and heightmap conditions to drive the low-level physical control policy. (c) The multi-condition physics policy. We divide common skills into 3 categories: Lococmotion, HSI, and DOI. Skills in the same category share similar task observations and reward computations.}
     \label{fig:pipeline}
\end{figure*}

\section{Method}
We present \ours as a hierarchical character animation system that leverages LLMs for high-level long-term script planning, multi-condition policies for low-level character control, and a finite state machine to bridge two levels. In \cref{subsec:3.1}, we first describe the construction of short script databases. \cref{subsec:3.2} then describes the generation of stylized long-term scripts using Retrieval-Augmented Script Generation (RASG). Finally,  \cref{subsec:3.3} explains the training of multi-condition policies and their scheduling through the finite state machine based on key frames. The supplementary material demonstrates our system's extensibility in adding new scene interaction skills.

\subsection{Short Script Database Construction}
\label{subsec:3.1}
A short script $p$ consists of a sequence of key frames $\{f_0, f_1, ..., f_N\}$. Each key frame $f = (s, o, c, e)$ specifies (1) a skill $s$ to execute, (2) a target object $o$ to interact with, (3) captions $c$ that describes motion attributes, and (4) the emotion or style  $e$ the motion expresses. Inspired by filmmaking, the short script uses only a few key frames to represent a short daily human-scene interaction segment. We add a concise one-sentence summary $u$ that encapsulates the core style or emotion and interaction events of the short script. We further separate the style or emotion keyword as a distinctive label $d$, as a conclusion of the keyframe style labels. hus, the final formation of the short script is $p = [\{ f_0, f_1, ..., f_N \}, u, d]$, serving as the foundational building block in the database. We prompt a Large Language Model (LLM)~\cite{gpt4} to generate a wide range of short scripts by providing it with the available skills, text captions, specific styles, and available objects. The LLM is tasked not only with creating coherent and lifelike key frame sequences but also with generating matching summaries $u$. These short scripts are further categorized based on their distinct emotion or style labels for better modular organization.
To enable retrieval, we employ CLIP~\cite{CLIP} to extract embeddings from the summaries of the short scripts. The extracted embeddings act as keys for efficient and precise retrieval within the database.

\subsection{Retrieval Augmented Script Generation}  
\label{subsec:3.2}  
Long-term script generation with LLMs faces challenges such as redundancy, lack of diversity, and insufficient guidance in maintaining coherent narratives. Previous works, such as \cite{unihsi}, focus on generating limited keyframes with minimal diversity, which constrains their ability to create engaging and robust long-term stories. Inspired by Retrieval-Augmented Generation (RAG)~\cite{RAG}, we propose a novel Retrieval-Augmented Script Generation (RASG) method to address these issues.  

To enhance long-term script generation, the LLM retrieves and builds upon the pre-generated short scripts based on user themes in the following steps:  

1) The LLM identifies \textit{M} styles most relevant to the theme, narrowing down the potential scope of retrieval. 
2) Semantic Similarity Retrieval: The user-provided theme sentence is extracted as a CLIP feature, which serves as the retrieval query. By computing the cosine distance between query and keys, the LLM retrieves top-\textit{k} of short scripts for each style. Resulting in \textit{M} × \textit{k} summaries being retrieved for further processing.  
3) Summary Filtering and Long Script Creation: The retrieved summaries are passed to the LLM. Then, based on the given scene layout, the LLM selects and combines suitable summaries into a cohesive narrative by logically concatenating keyframes.

To ensure executable permutations, we structure skills into tuples, such as \textit{(sit, getup)}, \textit{(lie, getup)}, \textit{(idle)}, \textit{(walk, carry)}, \textit{(walk, reach)}, etc. Notably, the \textit{walk} skill can serve as a transition motion between any skill tuples, enabling seamless connections across sequences. We use this rule to process the generated keyframes and add transitions for interaction skills.

\subsection{Multi-Condition Controller}
\label{subsec:3.3}
\noindent\paragraph{Overview} Once a long-term script generated, our goal is to direct a simulated character to perform the key frame sequence in complex 3D scenes. To train characters to complete tasks in a lifelike and stylized manner, we adopt a goal-conditioned RL framework with a text-conditioned discriminator~\cite{amp}. At each time step $t$, the policy $\pi(\rva_t | \rvs_t, \rvh_t, \rvg_t, \rvz)$ receives the humanoid proprioception $\rvs_t \in \mathcal{S}$, an egocentric heightmap $\rvh_t \in \mathcal{H}$, a task-specific goal state $\rvg_t \in \mathcal{G}$, and a language embedding $\rvz \in \mathcal{Z}$. The goal $\rvg_t$ specifies high-level task objectives that the character should achieve, such as contacting with a certain furniture or moving an object to a certain coordinate. The $\rvh_t$ is the egocentric heightmap around the character, representing the surrounding geometries. The language embedding $\rvz$ specifies the style that the character should use to achieve the desired task, such as walking excitedly or sitting with legs crossed. The policy $\pi$ then
samples an action $\mathbf{a}_t \in \mathcal{A}$. Applying the action $\mathbf{a}_t$, the environment performs state transition and the policy receives a reward $r_t$.
The objective is to learn a policy that maximizes the expected discounted return $J(\pi) = \mathbb{E}_{p(\tau | \pi)}\left[ \sum_{t=0}^{T-1}\gamma^{t} r_{t} \right]$, where $T$ is the horizontal length and $\gamma \in [0, 1]$ defines the discount factor. In order to train the policy $\pi$ to perform the task using diverse motion styles, we utilize a reward function consisting of two components:
$    r_t = \lambda^\text{style} r^\text{style}_t + \lambda^\text{task} r^\text{task}_t,$
where $r^\text{style}_t$ is a style reward modeled by the text-conditioned motion discriminator, and $r^\text{task}_t$ is a task-specific reward with coefficient $\lambda^\text{task}$.

\noindent\paragraph{Finite State Machine}
\label{subsec:3.4}
As illustrated in Fig~\ref{fig:pipeline}, our framework integrates several reusable policies, serving as low-level controllers. We have trained 7 policies: the \walk policy $\pi_w$, \idle policy $\pi_i$, \sit policy $\pi_s$, \lie policy $\pi_l$, \reach policy $\pi_r$, \getup policy $\pi_g$ and \carry policy $\pi_c$.

Following~\cite{interscene}, the FSM determines when to transition between skills. For instance, it initiates the next skill when the overlap time between the character's root and its target position exceeds a specific threshold. This simple rule-based FSM allows users to achieve desired long-term human motions in complex 3D scenes. Compared to the recent work InterScene~\cite{interscene}, our FSM contains egocentric heightmaps by frame and text embedding by skill, which could ensure scene understanding and semantic control.

\noindent\paragraph{Language Condition}
To control policy language constraints, we build an embedding space where motion representations are aligned with natural language descriptions. Given a motion clip $\hat{\mathbf{m}} = (\hat{\mathbf{q}}_1, \ldots, \hat{\mathbf{q}}_n)$, the motion encoder $\rvz = \text{Enc}_m(\hat{\mathbf{m}})$ maps the motion to a unit sphere embedding $\|\rvz\| = 1$, while corresponding text captions are processed through a pre-trained CLIP~\cite{CLIP} encoder $\text{Enc}_l$ and use fully connected layers to match the latent dimensionality. The training combines reconstruction and alignment losses to ensure that motion and text embeddings effectively correspond to each other. For further details on the network architecture and training losses, please refer to the Supp. Mat.

\noindent\paragraph{Scene Condition}
To enhance the humanoid's navigation and interaction capabilities, it is crucial to maintain environmental awareness to prevent collisions. We draw inspiration from methods such as \cite{wang2022towards,won2022physics, nsm, unihsi}, which utilize environmental sampling for humanoid observations. A square, ego-centric heightmap is generated to capture the elevation of surrounding objects. See in \cref{fig:pipeline}. Consistent with UniHSI~\cite{unihsi}, we pre-generate pointclouds for each scene. However, creating detailed pointclouds while preserving surface intricacies is computationally intensive.
To enhance the humanoid's understanding of complex surfaces for sitting or lying, we pre-generate scene pointclouds by voxelizing the objects within the bounding box range. The egocentric heightmap is updated by calculating the nearest object's pointclouds only when the object is sufficiently close to the humanoid's root position. The heightmap is a 12$\times$12 grid with an adjacent distance of 0.15 meters. We flatten the heightmap grid to a vector and concatenate it into the observation.

\noindent\paragraph{Universal Goal Condition} We consider 7 distinct scene interaction skills. To reduce the development overhead of diverse task-specific configurations, we implement all interaction tasks based on 3 task templates: Loco~(\walk and \idle), HSI~(\sit\, \lie, \reach and \getup) and DOI~(\carry). The implementation details are as follows:
\begin{itemize}
    \item \textbf{Loco tasks} require the humanoid to position its pelvis at a target 2D location $\rvg \in \mathbb{R}^2$ . For \walk, the location is set $\geq 1m$ from the humanoid's initial position, whereas the location of \idle is identical to the humanoid's current position, encouraging pacing in place.
    
    \item \textbf{HSI tasks} require a specific body joint to contact with the surface of a target object. We constrain the pelvis joint in \sit, \lie, and \getup, and use either the left or right hand for \reach. The target location $\rvg \in \mathbb{R}^3$ is determined by the nearest 3D point on the object's interactable surface.

    \item \textbf{DOI tasks} no longer constrain body joints, but encourage the character to move the dynamic object's root to a target 3D location. We use the bounding box coordinates of the object $\rvg^{bbox} \in \mathbb{R}^{3 \times 8}$ and the target location $\rvg^{tar} \in \mathbb{R}^3$ as the goal condition $\rvg = \{ \rvg^{bbox}, \rvg^{tar}  \}$.

\end{itemize}

\noindent Using sparse goal conditions can effectively train policies to perform scene interaction tasks~\cite{interphysics,interscene,coohoi}. However, we cannot control motion styles via these conditions. Tracking-based methods~\cite{unihsi,luo2023perpetual,tessler2024maskedmimic,xu2025intermimic} enable fine-grained control of each frame but require accurate stylized reference motions as dense input conditions. We employ a conditional discriminator~\cite{case,tessler2023calm} to inject text-based style control into policies. Unlike motion~\cite{tessler2023calm} or one-hot ~\cite{case} conditions, language is a more intuitive interface for LLMs and users.

\noindent\paragraph{Policy Training}

 We train 7 task-specific policies: (1) \walk, (2) Idle, (3) \sit, (4) \lie, (5) \reach, (6) \getup, and (7) \carry. We provide \walk, \idle, \sit, \lie, \carry policies with text conditions since these behaviors contain diverse interaction styles that represent vivid emotions. For \reach and \getup, we do not use text conditions. 
\begin{itemize}
    \item \textbf{Initialization.} Following UniHSI~\cite{unihsi}, we create the environment by randomly sampling objects from 3DFront~\cite{3dfront}. For HSI skills, we initialize characters using reference state initialization~\cite{deepmimic} and default pose initialization with a random global rotation and location\cite{unihsi, interscene} nearby the object. For locomotion skills, we randomly sampled on the whole ground plane while calculating the collision with the objects. For DOI skills, we randomly sample target position on the whole ground plane, and initialize objects in the humanoid's hands from reference object motion. Notebly, we add \walk motion data to the initiate reference state data during the training of all the skills because we use \walk as the transition between different interactions.
    
    \item \textbf{Rewards.} See the detailed reward function in Supp.Mat. 

    \item \textbf{Reset and early termination conditions.} Following \cite{amp}, we use a fixed episode length and fall detection as early termination triggers. We also use early termination when the task is accomplished for a certain time~\cite{interscene} or the contact forces are extremely large~\cite{unihsi}.
    
\end{itemize}

\section{Experiments}
\label{sec:exp}

\subsection{Dataset}

\begin{table}
\centering
\resizebox{0.95\linewidth}{!}{%
\begin{tabular}{c|cc|cccc|c}
\toprule[1.5pt]
\multirow{2}{*}{Datasets} & \multicolumn{2}{c|}{Loco}        & \multicolumn{4}{c|}{HSI}                                                                 & DOI   \\ 
 & Walk & Idle & Sit & Lie & Getup & Reach & Carry \\ \midrule
SAMP~\cite{samp}                      &    \fbox{20.6} &  -    &  35.2  &   14.8 &    \fbox{11.2}  &    -   &      - \\ 
COUCH~\cite{couch}                   & -    &   -   &   36.4 & -   &   \fbox{23.4}   &     -  & -      \\ 
Circles~\cite{circle}                   &  -   &    -  &  -  &  -  &   -   &    \fbox{3.6}   & -      \\ 
100Style~\cite{100style}                  & 203.1    &   -   &  -  &  -  &    -  &  -     & -      \\ 
AMASS~\cite{amass}                  & \fbox{8.2}   &    -  &   - &  -  &  -    &  -     &  \fbox{3.4}     \\ \midrule
ViconStyle                &   -  &   12.0   & -   &   21.9 &   \fbox{11.7}   &    -   &   26.0    \\ \bottomrule[1.5pt]
\end{tabular}}
\vspace{-8pt}
\caption{Mixture of collected stylized motion datasets.}
\vspace{-10pt}
\label{tab:data_mix}
\end{table}

We show our collected mixture of 6 motion dataset in \cref{tab:data_mix}. We show the skill for training and the motion duration in minutes. The number with black bounding-box like \fbox{20.6}, means the 20.6 minutes of motion in this dataset do not have style diversity, only counted as \textit{neutral}. ViconStyle is our captured dataset, which supplements for the quantity and the category of stylized motions. See details in Supp.Mat. We annotate all the motion clips with captions and style labels. For each caption, we provide 5 synonymous sentences with the help of LLM~\cite{gpt4}.
Besides neutral, we categorize the emotion or style of the remaining motions into 8 categories: \textit{happy, angry, hurried, tired, sad, stressed, drunk, and relaxed.} We left-right-flip all the motions so we get double the amount, and the captions are flipped concerning body joint symmetry as well.

For 3D objects, we use the furniture and scene layouts from the 3DFront~\cite{3dfront} dataset for training. Since 3DFront does not provide segmentation information, we voxelize the object meshes and segment the point clouds based on normal vectors to get the affordance surface.

\subsection{Motion Metrics}
To evaluate motion diversity, we use two metrics from the previous papers: Fréchet Inception Distance (FID)~\cite{mdm, case} and Average Pairwise Distance (APD)~\cite{case,wang2022towards}. FID measures the similarity between the distributions of generated and real data in a feature space, reflecting the realism and quality of the generated motions. Lower FID values indicate closer alignment with real data.
APD, on the other hand, quantifies the diversity within the generated motions by calculating the average pairwise distance between samples. Higher APD values indicate greater diversity in the generated motions. We calculate FID and APD on joint rotations and positions.

We follow \cite{samp, unihsi} that uses \emph{Success Rate} and \emph{Contact Error} as the main metrics to measure the quality of interactions quantitatively. Success Rate records the percentage of trials that humanoids successfully complete the contact within a certain threshold. We follow \cite{unihsi, interscene, interphysics} to set the threshold of \sit as 20cm, \reach as 20cm, \lie as 30cm, \carry as 20cm.

To evaluate the generation quality of long-term scripts, we also involve user study and SBERT~\cite{sbert} Model, please see the metrics in the corresponding part.

\subsection{Comparison with SOTA methods}
\subsubsection{Physical Performance for Different Skills}
Our method achieves better or comparable results across various metrics in \cref{tab:sota}. Unlike previous physics-based methods~\cite{unihsi, interscene, interphysics} which only care about contact but not styles, our result is achieved on 4096 random text conditions sampled from the datasets. The previous methods could be viewed as just a specific situation of our model. Under this background, we can see from \cref{tab:sota} that our results are only slightly lower than the best methods in \reach and \carry skill. Since Interphys~\cite{interphysics} have not released their code and carry motion data, we only train on the small amount of carry motion in AMASS~\cite{amass} for \cref{tab:sota}.

\begin{table}[t]
  \begin{center}
  
\resizebox{\linewidth}{!}{%
    \begin{tabular}{l|c|c|c|c|c|c|c|c}
    \toprule[1.5pt]
    \multirow{2}*{Methods}
    & \multicolumn{4}{c|}{Success Rate (\%) $\uparrow$}
    & \multicolumn{4}{c}{Contact Error $\downarrow$}\\
    & Sit & Lie & Reach & Carry& Sit & Lie & Reach & Carry\\
    \midrule
    InterPhys  \citep{interphysics} & 93.7 & 80.0 & - & 94.3 & 0.09 & 0.30 & - & \textbf{0.08} \\

    InterScene~\cite{interscene} & 97.8 &-  &  -& - & 0.04   & - &- & -\\
    UniHSI\cite{unihsi}     & 94.3 & {81.5} & \textbf{97.5} & - & \underline{0.032} & 0.061 & \textbf{0.016} & -  \\ \midrule
    SIMS & \underline{98.1} & \underline{87.6} &  \underline{95.2} & 92.9  & \textbf{0.028} & \underline{0.049} & \underline{0.026} & 0.099 \\
    SIMS~(+data) & \textbf{98.4} & \textbf{89.6} & -  & \textbf{96.4}  & 0.033 & \textbf{0.048} & - & \underline{0.085} \\
    \bottomrule[1.5pt]
    \end{tabular}}
\vspace{-8pt}
      \caption{Comparision on Baseline Models. For fair comparison, our \sit, \lie, and \reach policies are only trained on SAMP~\cite{samp} here. While our \carry policy is trained on the small amount of carry motions from AMASS~\cite{amass}. (+data) here represents our results trained on available motions from the mixture of 6 datasets.}
      \label{tab:sota} 
  \end{center}
  \vspace{-10pt}
\end{table}

\begin{figure*}[th]
    \centering
    \includegraphics[width=0.99\linewidth]{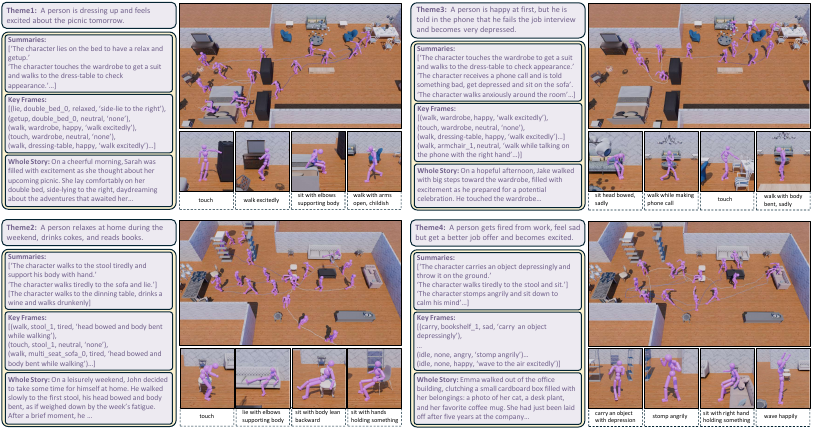}
    \caption{Long-term scripts with detailed keyframes and vivid final stories in two complex 3D scenes generated by our complete system. Upper: character in the bedroom and living room. Lower: character in the living room, dining room, and study room. We briefly demonstrate the retrieved summaries, key frames and part of the final long stories.}
    \label{fig:qualitive}
    
\end{figure*}

\subsubsection{Motion Diversity for Different Skills}
We compare motion diversity in the \sit and \lie skills with UniHSI~\cite{unihsi} and our re-implemented Interphys~\cite{interphysics}. All experiments are conducted on a single RTX 4090 GPU, running 1024 sequences and aggregating the results over 10 trials. For each sequence, the text condition is randomly sampled from the dataset. To test UniHSI~\cite{unihsi}, we randomly sample contact pairs from the provided chain of contacts from the generated ScenePlan dataset. We measure the FID between the generated motions and that of reference motions from SAMP~\cite{samp}. The APD measures the diversity among the generated motion sequences. As shown in \cref{tab:fid_apd}, our results significantly outperform UniHSI in both FID and APD metrics. Our method achieves lower FID, indicating motions produced from ours are closer to the
distribution of reference motions. Notably, the APD results highlight that the motions generated by UniHSI are nearly identical, demonstrating a lack of diversity. Our method also surpass the re-implemented InterPhys~\cite{interphysics}.
\begin{table}
\centering
\resizebox{\linewidth}{!}{%
\begin{tabular}{C{2cm}|C{1cm}|C{1cm}|C{1cm}|C{1.6cm}|C{1.6cm}|C{1.6cm}}
\toprule[1.5pt]
\multirow{2}{*}{Method} & \multicolumn{3}{c|}{FID$\downarrow$}                                   & \multicolumn{3}{c}{APD$\uparrow$}                                   \\ 
 & Sit & Lie  & Carry & Sit & Lie  & Carry \\ \midrule

InterPhys*~\cite{interphysics}
& - &- &  81.0  & - & - &12.41$\pm$0.19 \\
UniHSI~\cite{unihsi}                      &  153.84   &    211.22 &  -  & 1.14$\pm$0.01   & 
1.35$\pm$0.02 & - \\
\midrule
SIMS                   &  \textbf{125.66}    &  \textbf{171.24}  &  \textbf{65.14} & \textbf{16.55$\pm$0.54}  &   \textbf{16.40$\pm$0.94} & \textbf{14.36$\pm$0.12}\\ \bottomrule[1.5pt]

\end{tabular}}
\caption{Motion diversity results. InterPhys~\cite{interphysics} is not released, so we report our re-implemented version here. For fair comparison, our \sit, \lie, and \reach policies are only trained on SAMP~\cite{samp} here. While the \carry policy and the re-implemented InterPhys are both trained on the carry motions from ViconStyle. }
\label{tab:fid_apd}
\vspace{-10pt}
\end{table}

\subsubsection{User Study on SOTA Long-Term HSI Methods}
To further evaluate the control capabilities of the long-term scripts, we conducted a user study on the rendered videos generated from different methods. We use the same category of interactions to drive the characters in the scenes. 30 participants were asked to rate the physical realism, motion diversity, split engagement and emotion resonace of the videos produced by each method on a scale from 1 (poor) to 5 (excellent). In ~\cref{tab:user_study}, the results indicate that our approach significantly outperformed UniHSI, demonstrating its effectiveness in both body motion superiority and script superiority in the generated animations. 

\begin{table}[hbt]
\centering
\resizebox{0.8\linewidth}{!}{%

\begin{tabular}{ll|C{2cm}|C{2cm}}
\toprule[1.5pt]
\multicolumn{2}{l|}{Metrics}                                      & UniHSI  & SIMS \\ \midrule
\multicolumn{1}{l|}{\multirow{2}{*}{Motion}} & Physical Realism~$\uparrow$  &  2.6
      &  \textbf{3.4}
    \\ 
\multicolumn{1}{l|}{}                        & Motion Diversity~$\uparrow$   &    2.9
    &     \textbf{3.6}
 \\ \midrule
\multicolumn{1}{l|}{\multirow{2}{*}{Script}} & Plot Engagement~$\uparrow$    &    2.4    & \textbf{3.0}
     \\
\multicolumn{1}{l|}{}                        & Emotional Resonace~$\uparrow$ &    3.0    &    \textbf{3.8}
  \\ \bottomrule[1.5pt]
\end{tabular}
}
\vspace{-3pt}
\caption{User Study on SOTA long-term HSI methods. \ours outperforms the SOTA method UniHSI by a significant margin.}
\label{tab:user_study}
\vspace{-3pt}
\end{table}

\begin{figure}
    \centering
       \begin{center}
    \includegraphics[width=0.96\linewidth]{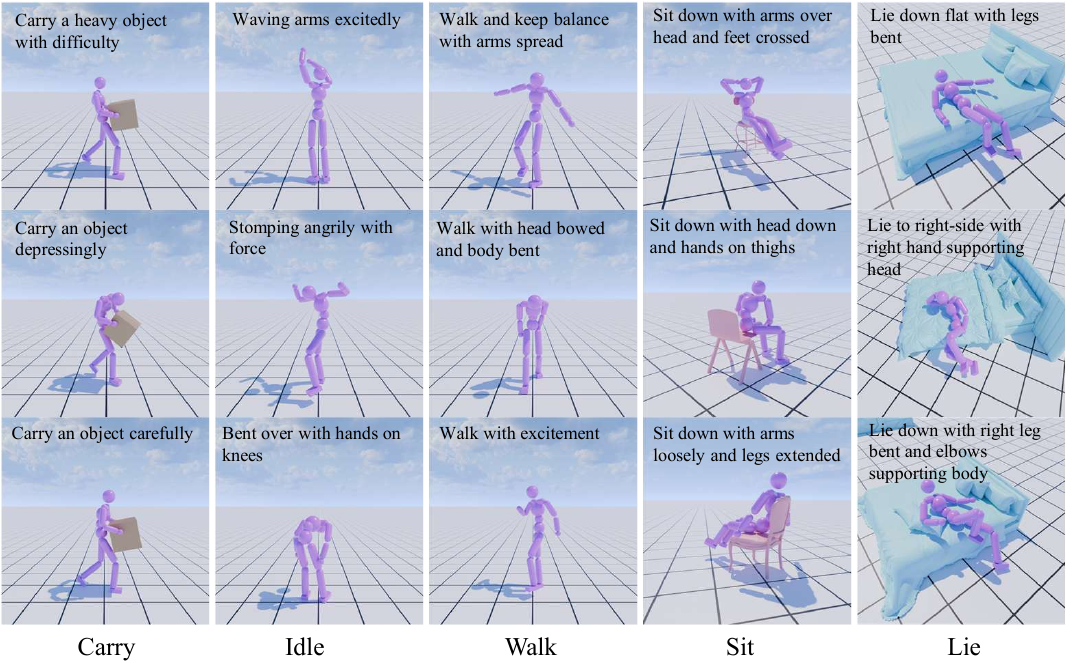}
    \end{center}
    \vspace{-10pt}
    \caption{Qualitative results for skills with different text conditions.}
    \label{fig:skills}
\end{figure}

\subsection{Ablation Study on SIMS}
\begin{table}
\centering
\resizebox{0.8\linewidth}{!}{
\begin{tabular}{l|c|c}
\toprule[1.5pt]
Method & SBERT Similarity~\cite{sbert}$\downarrow$ & Average Generation Time(s)$\downarrow$ \\ \midrule
LLM    &        0.8167           &  12.2               \\ 
RASG    &        \textbf{0.7759}           &     \textbf{7.32}            \\ \bottomrule[1.5pt]
\end{tabular}}
\caption{Ablation on script generation methods.}
\label{abl_rag}
\end{table}
\subsubsection{Direct Generation   \textit{vs.} RASG.}
We compare our RASG method with direct LLM generation using GPT-4~\cite{gpt4}. For direct LLM generation, we provide the LLM with all the available skills as input. To evaluate the narrative diversity and generation efficiency of our approach, we measure the cosine similarity of SBERT~\cite{sbert} embeddings and the generation time. Our method achieves lower cosine similarity among the generated stories, indicating that it produces more diverse scripts. For generation time, we require the LLM to generate approximately 20 keyframes for direct generation method. For the RASG method, we ask LLM to retrieve 4-5 short scripts, which are approximately 20 keyframes in total. The results are evaluated on 200 generated samples separately.

\subsubsection{Generalization on Unseen Objects}
\begin{table}
\centering
\resizebox{0.7\linewidth}{!}{
\begin{tabular}{c|cc|cc}
\toprule[1.5pt]
\multirow{2}{*}{Datasets} & \multicolumn{2}{c|}{Success Rate(\%)$\uparrow$}  & \multicolumn{2}{c}{Contact Error$\downarrow$}             \\  
      & Sit   & Lie  & Sit   & Lie  \\ \midrule 
PartNet~\cite{partnet}                        &   98.7   &    87.6       & 0.028  & 0.065     \\ 
3DFront~\cite{3dfront}                    &     96.9 &   89.7   &   0.014    &  0.030      \\ \bottomrule[1.5pt]
\end{tabular}

}
\caption{Results on PartNet and 3DFront. The policies are trained on 3DFront's furniture only.}
\label{abl_scene}
\vspace{-10pt}
\end{table}
In \cref{abl_scene}, we show the physical performance of interaction skills on PartNet~\cite{partnet} and 3DFront~\cite{3dfront}. Note that our policies are only trained on the objects from 3DFront. From the table, we can see our results could achieve as good performance on unseen objects, mainly due to the generalization ability of heightmap design.

\subsubsection{Scale Up on New Motion Datasets}

To prove the reliable of the proposed datasets, and the generality of our text-conditioned policy, we report the Success Rate and APD for \walk, \carry, \sit, and \lie skills in \cref{abl_walk_data}, \cref{abl_hoi_data}, and \cref{abl_hsi_data}. From the tables, we could find that with more data, \walk achieves a higher success rate mainly because AMASS provides stable neutral walking and running motions. The APD changes little because 100Style also contains neutral walking styles. For carry skill, since ViconStyle is the first dataset containing stylized carrying motion, both metrics increase by a large margin.  For HSI skills, sit and lie both become slightly better with the introduction of COUCH and ViconStyle dataset. Couch provides more stylized sitting motions and ViconStyle provides more stylized lying motions.

\begin{table}[ht]
\centering
\begin{minipage}{0.46\linewidth}
\centering
\resizebox{\linewidth}{!}{%
\begin{tabular}{c|c|c}
\toprule[1.5pt]
\multirow{2}{*}{Datasets} & Success Rate(\%)$\uparrow$ & {APD$\uparrow$} \\ 
                          & Walk         & Walk  \\ \midrule
100S     &     92.6         &   14.83$\pm$0.35                \\ 
A+100S                  &   95.1     & 14.88$\pm$0.29                               \\ \bottomrule[1.5pt]
\end{tabular}}
\vspace{-10pt}
\caption{Dataset ablation on \walk Skill. 100S: 100Style
, A: AMASS.}
\label{abl_walk_data}
\end{minipage}
\hfill
\begin{minipage}{0.46\linewidth}
\centering
\resizebox{\linewidth}{!}{%
\begin{tabular}{c|c|c}
\toprule[1.5pt]
\multirow{2}{*}{Datasets} & Success Rate(\%)$\uparrow$ & APD$\uparrow$ \\ 
                          & carry         & carry  \\ \midrule
A                  & 92.9  &   14.36$\pm$0.12                                                                \\ 
A+VS              &       96.4       & 14.92$\pm$0.23                                                        \\ \bottomrule[1.5pt]
\end{tabular}}
\vspace{-10pt}
\caption{Dataset ablation on \carry Skill. A: AMASS. VS: ViconStyle.}
\label{abl_hoi_data}
\end{minipage}
\vspace{-10pt}
\end{table}
\begin{table}
\centering
\resizebox{0.9\linewidth}{!}{%
\begin{tabular}{c|cc|cc|cc}
\toprule[1.5pt]
\multirow{2}{*}{Datasets} & \multicolumn{2}{c|}{Success Rate(\%)$\uparrow$} & \multicolumn{2}{c|}{Contact error$\downarrow$} & \multicolumn{2}{c}{APD$\uparrow$}          \\  
                          & Sit   & Lie  & Sit   & Lie   & Sit & Lie  \\ \midrule 
S    &   95.5   &   86.9   &   0.040   &  0.055     &    16.43$\pm$0.90 &   16.40$\pm$0.94  \\ 
S+C    &     96.9   &   -   &  0.014    &    -   & 16.52$\pm$0.47 &    -   \\ 
S+C+VS                   &    -  &    89.7  &  -    &   0.030    &  -  &   16.84$\pm$1.28   \\ \bottomrule[1.5pt]
\end{tabular}}
\vspace{-3pt}
\caption{Dataset ablation on HSI Skills. S: SAMP~\cite{samp}, C: Couch~\cite{couch}, VS: ViconStyle}
\label{abl_hsi_data}
\vspace{-3pt}
\end{table}

\subsubsection{Ablation of Policy Settings}
We conducted an ablation study on different settings of our control policy, comparing the \emph{Success Rate} and \emph{Contact Error} for variations without heightmap and without text embedding. Both variants showed degraded performance. The height map provides essential information about the surrounding environment so the performance becomes worse when interacting with objects. When trained without text embedding, the APD metric shows an obvious degradation.  
\begin{table}
\centering
\resizebox{0.9\linewidth}{!}{%
\begin{tabular}{c|ccc|ccc}
\toprule[1.5pt]
\multirow{2}{*}{Setting} & \multicolumn{3}{c|}{Success Rate(\%)$\uparrow$}                                                       & \multicolumn{3}{c}{APD$\uparrow$}                             \\ 
& Sit & Lie & Carry & Sit & Lie  & Carry   \\ \midrule
w/o text &  89.7   &  89.6   &   92.4   &   16.29$\pm$0.22    &  16.59$\pm$0.28   &   12.41$\pm$0.19      \\ 
 w/o htmp  &   88.7  &  79.8   &  -       &  16.18$\pm$0.19 &  16.94$\pm$0.29   &     -     \\ 
 SIMS(ours)   &    96.9 &   89.7     &    96.4   & 16.52$\pm$0.47    &  16.99$\pm$1.28   &  14.92$\pm$0.23           \\ \bottomrule[1.5pt]
\end{tabular}}
\caption{Ablation on different policy settings.}
\label{abl_policy}
\vspace{-8pt}
\end{table}

\subsection{Qualitative Results}
We show 4 generated long narratives executed by our policies in two large indoor scenes. The details can be viewed in \cref{fig:qualitive}. In \cref{fig:skills}, we also showed some qualitative samples for 5 skills: \carry, \idle, \walk, \sit, and \lie. We suggest the readers to refer to the demonstration videos for a better knowledge of our ability to generate long-term stylized motions.
\section{Conclusion}

In this paper, we analyze and compare the current advancements in long-term human-scene interaction tasks, highlighting the lack of generating animations that are both physically plausible and stylistically expressive. To address this, we propose a novel framework for synthesizing long-term human-scene interactions by leveraging Retrieval-Augmented Generation as high-level planners and a multi-condition control policy as the low-level controller. By incorporating both stylized script generation and a stylized control policy, our approach facilitates the creation of diverse, expressive, and physically coherent long-term animations. Furthermore, the processed datasets open up new possibilities and directions for future research in this field.
\section{Furture Work}

In the future, it will be essential to collect more human motion data that captures realistic emotions and diverse styles. Additionally, exploring humanoid models with articulated fingers presents a promising avenue for research. Introducing multi-agent in HSI could also broaden the possibilities for physical animations.
\clearpage


\clearpage
\setcounter{page}{1}
\maketitlesupplementary

\begin{figure*}
    \centering
    \includegraphics[width=0.98\linewidth]{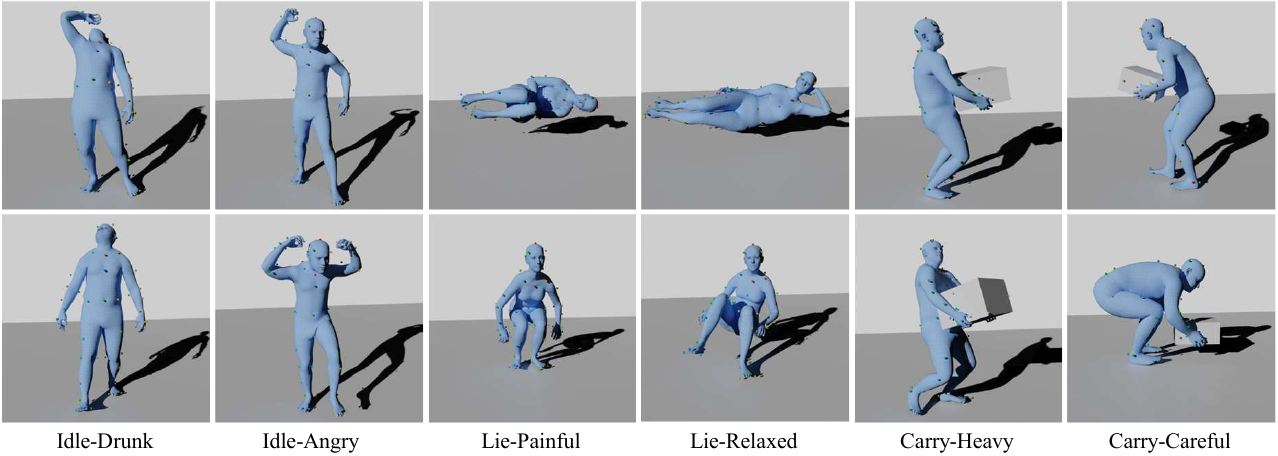}
    \caption{ViconStyle demos.}
    \label{fig:viconstyle}
\end{figure*}

\section{Reward Templates}
In this section, we introduce the reward functions in 3 parts: locomotion (Loco), human-scene interaction (HSI), and dynamic object interaction (DOI).
\begin{itemize}

\item \textbf{Loco Reward.} 
The locomotion reward is defined in Equation \ref{eq:walk_total_reward}. 
The overall reward comprises the far $r_t^{far}$, near $ r_t^{near}$, and standstill $ r_t^{still}$ rewards. The standstill reward ensures that the humanoid remains static once the target position has been reached.
Given a target position  $x^* $ of the character's root $x^{root} $, a target direction  $d^*_t $, and a target scalar velocity $g_t^{vel}$, the task reward is defined as:

\begin{equation}
r^G_t=\left\{
\begin{aligned}
0.4 \ r_t^{near} &+ 0.5 \ r_t^{far} + 0, \left \| x^* - x^{root}_t \right \|^2 > 0.5, \\
0.4 \ r_t^{near} &+ 0.5 + 0.1 \ r_t^{still}, \text{otherwise.}
\end{aligned}
\right.
\label{eq:walk_total_reward}
\end{equation}

\begin{equation}
\begin{aligned}
r_t^{far} &= 0.6 \ \text{exp}\big(-0.5 \left \| x^* - x_t^{root} \right \|^2 \big) \\ 
&+ 0.2 \ \text{exp} \big ( -2.0 \left \| g_t^{vel} - d_t^* \cdot \dot{x}^{root}_t \right \|^2 \big ) \\
&+ 0.2 \  \left \| d^*_t \cdot d^{facing}_t \right \|^2
\end{aligned}
\end{equation}

\begin{equation}
r_t^{near} = \text{exp}\big(-10.0 \left \| x^*  - x_t^{root} \right \|^2 \big)
\end{equation}

\begin{equation}
 r_t^{still} = \text{exp}\big(-2.0 \left \| \dot{x}^{root}_t - \dot{x}^{root}_{t-1}  \right \|^2)
\end{equation}
The main difference between \walk and \idle reward is that we allow a large distance threshold for \idle. We restrict the \walk skill to reach the target coordinate as close as possible, but only restrict \idle to maintain inside 3 meters distance.

\item \textbf{HSI Reward.} The HSI reward is defined in Eq~\ref{eq:sit_total_reward}. 
The far reward $r_t^{far}$ is to encourage the humanoid's pelvis $x^{root}$ to reach the target coordinate $x^*$ with the target speed $g_t^{vel}$ and target direction $d^*_t $. Like UniHSI~\cite{unihsi}, the near reward $r_t^{near}$ encourages the humanoid's certain joint to contact the nearest point in an interactable part $p$ of the target object. For \sit we require pelvis to contact the target sitting point, while for \lie we require pelvis to reach the nearest point on the bed's surface. For \reach, either left or right hand is supposed to reach the object's surface. The task reward is defined as:

\begin{equation}
    r^G_t=\left\{
    \begin{aligned}
    0.7 \ r_t^{near} &+ 0.3 \ r_t^{far}, \left \| x_t^* - x^{root}_t \right \|^2 > 0.5 \\
    0.7 \ r_t^{near} &+ 0.3, \text{otherwise}
    \end{aligned}
    \right.
\label{eq:sit_total_reward}
\end{equation}
\begin{equation}
    \begin{aligned}
    r_t^{far} &= \text{exp} \big ( -2.0 \left \| g_t^{vel} - d_t^* \cdot \dot{x}^{root}_t \right \|^2 \big )
    \end{aligned}
\label{eq:far_reward}
\end{equation}
\begin{equation}
r_t^{near} = \text{exp}\big(-10.0 \left \| x_t^{*} - x_t^{root} \right \|^2 \big)
\label{eq:near_reward}
\end{equation}

\textbf{Getup Reward.} 
The \getup skill is developed through step goals, which combine walk and contact rewards. If the contact goal has not been reached, the reward encourages the humanoid to sit or lie on the object. Conversely, when the contact goal is achieved, the reward motivates the humanoid to elevate its pelvis to a standing position. The formulation for this reward system aligns with that of the contact reward $r^{near}_t$.

\item \textbf{DOI Reward.} 
In this version, we only implement \carry skill in DOI task. However, our DOI reward could serve as a universal template for dynamic object interactions, like push, throw, etc. The reward is split into 3 parts: walk reward $r^{walk}_t$, encourages the humanoid walk to the object first; hand contact reward $r_t^{hand}$, encourages the humanoid place its hand on the object before the task been completed; moving reward $r_t^{carry}$, encourages to the object to the target position. 

\begin{equation}
r^G_t = \left\{
\begin{aligned}
&0.3 \ r_t^{walk} + 0.5 \ r_t^{carry}
 + 0.2 \ r_t^{hand}, && \|x_t^{obj} - x_t^{goal}\|^2 > 0.5, \\
&0.3 \ r_t^{walk} + 0.5 \ r_t^{carry} + 0.2, && \text{otherwise}.
\end{aligned}
\right.
\label{eq:doi_total_reward}
\end{equation}

\begin{equation}
\begin{aligned}
r_t^{walk} &=  0.8 \cdot \exp\big(- 10.0 \cdot \|x_t^{root} - x_t^{obj}\|^2\big) \\ 
           &+ 0.2 \cdot \exp\big(- 2.0 \cdot \|v_t^{root} - v_t^{goal}\|^2\big),
\end{aligned}
\end{equation}

\begin{equation}
\begin{aligned}
r_t^{hand} &= \exp\big(- 0.5 \cdot \|x_t^{hand} - x_t^{obj}\|^2\big) 
\end{aligned}
\end{equation}

\begin{equation}
\begin{aligned}
r_t^{carry} &=  0.7 \cdot \exp\big(- 10.0 \cdot \|x_t^{obj} - x_t^{goal}\|^2\big) \\ 
            &+ 0.3 \cdot \exp\big(- 2.0 \cdot \|v_t^{obj} - v_t^{goal}\|^2\big).
\end{aligned}
\end{equation}

\end{itemize}

\begin{figure}
    \centering
    \includegraphics[width=0.6\linewidth]{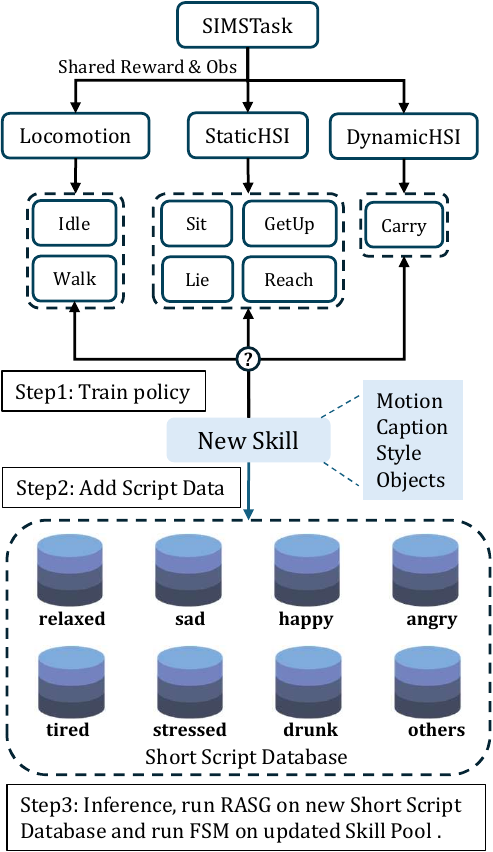}
    \caption{Scalability on new skills.}
    \label{fig:skill_scaleup}
\end{figure}

\section{Re-implemented MotionCLIP}

\begin{figure}
    \centering
    \includegraphics[width=0.8\linewidth]{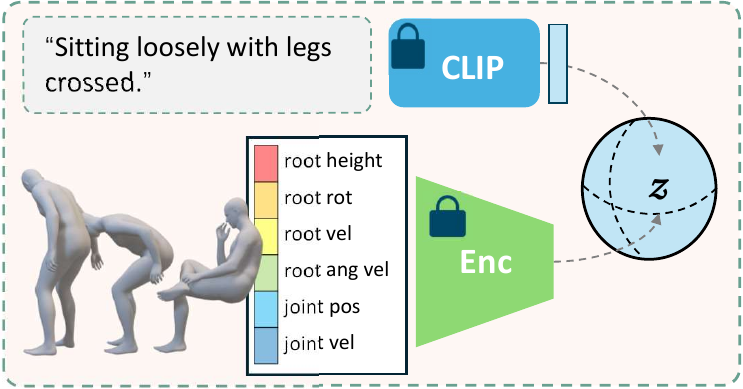}
    \caption{Our re-implemented MotionClip.}
    \label{fig:motionclip}
\end{figure}

To control the policy language constraints, we aim to construct an embedding space fed into the policy network, where the embedding aligns motion representation with their corresponding natural language descriptions. To do this, we follow \cite{padl, motionclip}, where a transformer auto-encoder is trained to encode motion sequences into a latent representation that aligns with the language embedding from a pre-trained CLIP text encoder \cite{CLIP}. Given a motion clip $\hat{\mathbf{m}} = (\hat{\mathbf{q}}_1, \ldots, \hat{\mathbf{q}}_n)$, a motion encoder $\rvz = \text{Enc}_m(\hat{\mathbf{m}})$ maps the motion to an embedding $\rvz$. The embedding is normalized to lie on a unit sphere $\|\rvz\| = 1$. We set the embedding size $\rvz$ to 64 to save the computation cost. For the text embedding, we first extract the feature with CLIP Encoder~\cite{CLIP} $\text{Enc}_l$ from caption $\rvc$, then use a multilayer perception $\text{MLP}_d$ to downsize the 512 dim CLIP feature to 64 dim and use an extra one $\text{MLP}_u$ to upsample it to 512 dim to maintain the semantic feature. The embedding $\rvz$ should be aligned with the downsized CLIP feature. See details in \cref{fig:motionclip} 
Following~\cite{motionclip}, $\text{Enc}_m\left(\mathbf{m} \right)$ is modeled by a bidirectional transformer \cite{bert2018}. The motion decoder is jointly trained with the encoder to produce a reconstruction sequence $\mathbf{m} = (\mathbf{q}_1, \ldots, \mathbf{q}_n)$ to recover $\hat{\mathbf{m}}$ from $\rvz$. The motion representation $\mathbf{q}$ we use is a set of character motion features, following the discriminator observation used in AMP~\cite{amp}. The auto-encoder is trained with the loss:
\begin{align}
\mathcal{L}_{\text{AE}} = \mathcal{L}_{\text{recon}}^{m} + \mathcal{L}_{\text{align}}^{m,t} + \mathcal{L}_{\text{recon}}^{t}.
\end{align}
The reconstruction loss $\mathcal{L}_{\text{recon}}^m$ measures the MSE error between the reconstructed sequence and original motion.

The alignment loss $\mathcal{L}_{\text{align}}^{m,t}$ measures the cosine distance between the motion embedding and the downsized CLIP feature:
\begin{align}
\mathcal{L}_{\text{align}}^{m,t} = 1 - d_{\text{cos}}\left(\text{Enc}_m\left(\hat{\mathbf{m}}\right), \text{MLP}_d(\text{Enc}_l(\rvc)\right)) .
\end{align}
The text embedding reconstruction loss $\mathcal{L}_{\text{recon}}^t$ measures the MSE distance between the reconstructed CLIP embedding and the original one:
\begin{align}
\mathcal{L}_{\text{recon}}^{t} = 
\| \text{MLP}_u(\text{MLP}_d(\text{Enc}_l(\rvc)))) - \text{Enc}_l(\rvc)
\|_2
\end{align}
The weights of $\text{Enc}_l$ are fixed during training. To maintain the semantic information, we follow the sampling strategy used in MotionCLIP \cite{motionclip}. We sample 300 frames from the 30fps motion data and use skip sampling for the motion clips that are longer than 10 seconds so that all the information is included.
\section{New Skill Scalability}
In \cref{fig:skill_scaleup}, we show the easy scalability of our framework. When new skills of new styles come, we need to train the corresponding skill based on the 3 kinds of templates, and expand the scripts database following the instruction of \cref{subsec:3.1}.

\section{ViconStyle Dataset}

We propose a comprehensive motion dataset called ViconStyle, in which well-labeled reconstructed motion clips with diverse styles and multiple skills are provided. 
\begin{figure}
    \centering
    \includegraphics[width=0.5\linewidth]{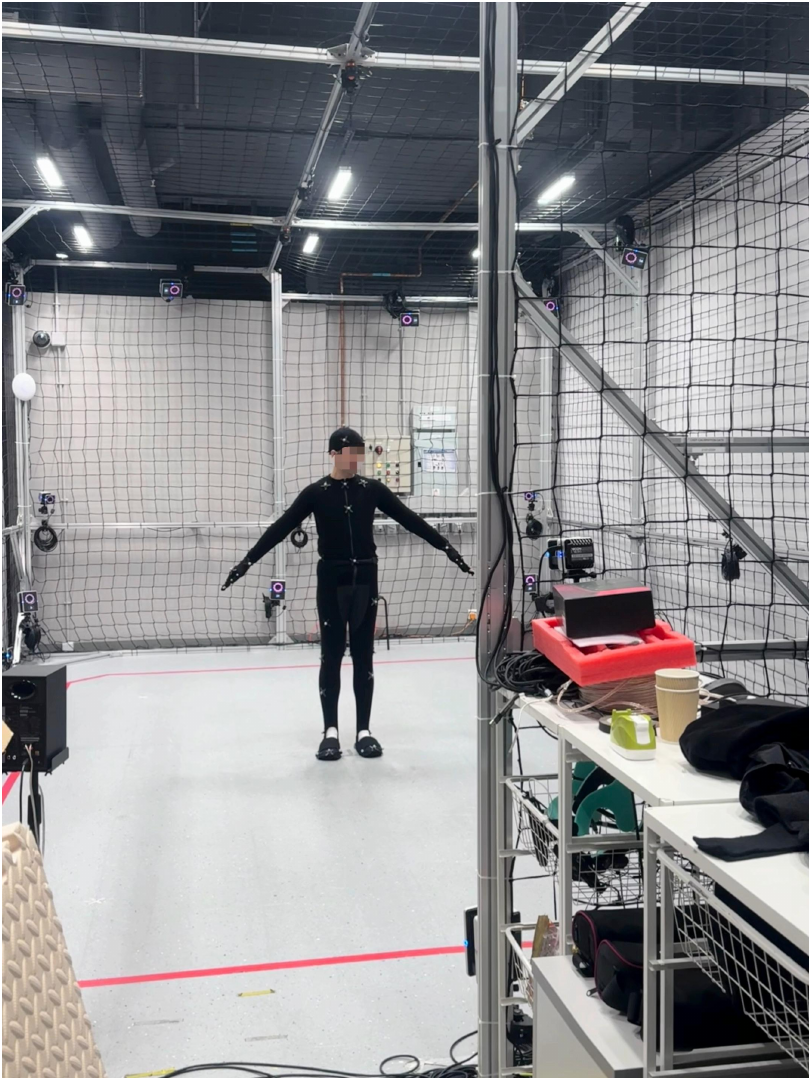}
    \caption{The motion capture environment of Vicon optical motion capture system.}
    \label{fig:mocap-setting}
\end{figure}

\subsection{Capture Setting}

The motion clips are captured with Vicon, an optical motion capture system, as shown in figure \ref{fig:mocap-setting}. All motion clips are captured with 120 fps. During the capture, we asked actors to interact with scene objects of different sizes and weights, such as lying on the sofa or carrying boxes. 

We used SOMA \cite{soma} to fit the SMPL \cite{smpl} body model and its pose parameters. The mocap data are then annotated with text descriptions containing motion details such as "hands on the thighs" and "lean back" and motion styles and emotions.

We also used a method to calculate the transformation and orientation and fit the size of the scene objects that we captured. We divide the reconstruction problem into two stages. In the first stage, we need to approximate the initial state of the scene objects. Since the scene objects are mainly boxes, the state estimation problem can be converted into an axis regression problem. We first regress the most suitable local coordinate by rotating the axis to minimum the max distance from the captured marker points to the axis. Then we move the origin point to the center of the bounding boxes of the marker points, and the scale can also be easily calculated. In the second stage, we trivially represent the subsequent transformation and orientation in the form of displacements and rotations relative to the initial frame. 

\subsection{Dataset Statistics}

We recruited three actors to capture the dataset. The motion clips we captured contain 7 skills and actors are asked to perform in different styles and add details in every motion clip. The motion data set is 71.6 minutes in length and has 415 clips in total. The information of the actors is listed in table \cref{tab:actor_information}, and the detailed statistics of the data set are listed in table \cref{tab:data_mix}.

\begin{table}
    \centering
    \resizebox{0.8\linewidth}{!}{
    \begin{tabular}{c|c|c|c|c}
    \toprule[1.5pt]
        Actors No. & Age & Gender & Height & Weight\\
        \midrule
        1 & 22 & Female & 168 & 55 \\
        
        2 & 22 & Male & 182 & 71 \\
        3 & 30 & Male & 175 & 85 \\
        
    \bottomrule[1.5pt]
    \end{tabular}}
    \caption{Actor information.}
    \label{tab:actor_information}
\end{table}

\subsection{Qualitative Results}

The captured motion contains diverse styles of \idle, \lie, \carry, and \getup skills. See ~\cref{fig:viconstyle} for demonstration.

\section{Short Script Examples}

\begin{table}[ht]
\small
\scalebox{0.76}{%
\begin{tabular}{p{0.16\linewidth}|p{0.16\linewidth}|p{0.16\linewidth}|p{0.54\linewidth}}\toprule[1pt]
\multicolumn{4}{l}{Summary: The character enjoys a \textbf{relaxed} afternoon in the living room.}\\ \midrule[1pt]

skill  &  style  & object &  captions   \\ \midrule[1pt]
loco & neutral & - & smoothly forward walk \\ \hline
idle  &  relaxed & - & relaxing body \\ \hline
sit  & relaxed &sofa & leaning back, legs straight, hands supporting head \\ \hline
getup&neutral & sofa  & -\\  \hline
touch & -  &  shelf & -
\\ \bottomrule
\end{tabular}}

\scalebox{0.76}{%
\begin{tabular}{p{0.16\linewidth}|p{0.16\linewidth}|p{0.16\linewidth}|p{0.54\linewidth}}\toprule[1pt]
\multicolumn{4}{l}{Summary: The character rushed \textbf{anxiously} through the living room.}\\ \midrule[1pt]

skill&style  & object &  captions   \\ \midrule[1pt]

loco & anxious & - & rush anxiously forward \\ \hline
touch & - & shelf & -  \\ \hline
idle & anxious & - & pace around nervously\\ \hline
loco & hurried & table & walk with large steps 
\\ \bottomrule
\end{tabular}}

\scalebox{0.76}{%
\begin{tabular}{p{0.16\linewidth}|p{0.16\linewidth}|p{0.16\linewidth}|p{0.54\linewidth}}\toprule[1pt]

\multicolumn{4}{l}{Summary: Character felt utterly \textbf{tired} and sleep in the bedroom.}\\ \midrule[1pt]
skill  &  style  & object &  captions   \\ \midrule[1pt]
idle & tired & - & bent over with hands on knees\\ \hline
loco & tired & lamp &  head bowed and body bent while walking\\ \hline
touch & - &lamp & - \\ \hline
loco & neutral & - &moving backward while walking \\ \hline
lie &tired & bed &lying down, legs straight
\\ \bottomrule
\end{tabular}}

\scalebox{0.76}{%
\begin{tabular}{p{0.16\linewidth}|p{0.16\linewidth}|p{0.16\linewidth}|p{0.54\linewidth}}\toprule[1pt]

\multicolumn{4}{l}{Summary: The character \textbf{happily} played and relaxed around the bedroom}\\ \midrule[1pt]
skill  &  style  & object &  captions   \\ \midrule[1pt]
loco & happy &   wardrobe&  excited walk \\ \hline
carry & happy & toy & carry object happily\\\hline
loco & happy &   sofa &  excited walk \\ \hline
sitdown & relaxed & sofa &  hands support body, cross-legged
\\ \bottomrule
\end{tabular}}

\scalebox{0.76}{%
\begin{tabular}{p{0.16\linewidth}|p{0.16\linewidth}|p{0.16\linewidth}|p{0.54\linewidth}}\toprule[1pt]
\multicolumn{4}{l}{Summary: The character is \textbf{angry} and knocks on the table, then sit.}\\ \midrule[1pt]
skill  &  style  & object &  captions   \\ \midrule[1pt]
loco & angry &  -&  angrily walking \\ \hline
idle & angry &  -& stomp angrily against the ground \\ \hline
touch & table & -\\\hline

sit &angry &   armchair & crossing arms
\\ \bottomrule
\end{tabular}}

\scalebox{0.76}{%
\begin{tabular}{p{0.16\linewidth}|p{0.16\linewidth}|p{0.16\linewidth}|p{0.54\linewidth}}\toprule[1pt]

\multicolumn{4}{l}{Summary: The character gets \textbf{drunk} and stumbles around the living room.}\\ \midrule[1pt]
skill  &  style  & object &  captions   \\ \midrule[1pt]
idle & drunk & - & stand drunkenly \\ \hline
loco & drunk &   sofa & walking drunkenly \\ \hline
sit &drunk & sofa & right leg held, left leg stretched out \\\hline
touch & sofa & - \\\hline
loco & drunk & sofa & walking drunkenly \\ \hline
lie &tired & sofa   & lying down, legs straight
\\ \bottomrule
\end{tabular}}

\scalebox{0.76}{%
\begin{tabular}{p{0.16\linewidth}|p{0.16\linewidth}|p{0.16\linewidth}|p{0.54\linewidth}}\toprule[1pt]

\multicolumn{4}{l}{Summary: The character feels \textbf{stressed} and seeks comfort in the living room.}\\ \midrule[1pt]
skill  &  style  & object &  captions   \\ \midrule[1pt]
sit &stressed & armchair & sitting with head bowed, hands resting on thighs \\ \hline
touch &   armchair & -\\ \hline
loco & stressed & sofa &walking slowly, hands behind back\\\hline
lie &stressed & sofa   & side-lie on left with left arm as pillow, legs bent
\\ \bottomrule
\end{tabular}}

\scalebox{0.76}{%
\begin{tabular}{p{0.16\linewidth}|p{0.16\linewidth}|p{0.16\linewidth}|p{0.54\linewidth}}\toprule[1pt]

\multicolumn{4}{l}{Summary: The character discovered an old vase on the shelf, settled on the sofa.}\\ \midrule[1pt]
skill  &  style  & object &  captions   \\ \midrule[1pt]
loco & neutral & & side-stepping\\ \hline
touch & neutral & shelf & - \\ \hline
carry & neutral & vase & carry object calmly \\ \hline
liedown & neutral & sofa & legs bend
\\ \bottomrule
\end{tabular}}

\caption{Examples in the Short Script Database.}
\label{tab:short_scripts}

\end{table}

We show some vivid examples in \cref{tab:short_scripts} for all the emotions/styles we use. Please check the skills, style label, object type, and captions, which are essential for FSM control.

\end{document}